# Selective De-noising of Sparse-Coloured Images

Arjun Chaudhuri
Indian Institute of Technology,Kharagpur

*Abstract*—Since time immemorial, noise has been a constant source of disturbance to the various entities known to mankind. Noise models of different kinds have been developed to study noise in more detailed fashion over the years. Image processing, particularly, has extensively implemented several algorithms to reduce noise in photographs and pictorial documents to alleviate the effect of noise. Images with sparse colours-lesser number of distinct colours in them-are common nowadays, especially in astronomy and astrophysics where black and white colours form the main components. Additive noise of Gaussian type is the most common form of noise to be studied and analysed in majority of communication channels, namely-satellite links, mobile base station to local cellular tower communication channel,et. al. Most of the time, we encounter images from astronomical sources being distorted with noise maximally as they travel long distance from telescopes in outer space to Earth. Considering Additive White Gaussian Noise(AWGN) to be the common noise in these long distance channels, this paper provides an insight and an algorithmic approach to pixel-specific de-noising of sparse-coloured images affected by AWGN. The paper concludes with some essential future avenues and applications of this de-noising method in industry and academia.

## I. INTRODUCTION

The understanding of the physics of noise and the knowledge of linear algebra has resolved the problem of distortion in image quality to a great extent. Researchers across the globe are involved in developing new mathematical models and investigating them to have a deeper conceptual understanding of noise at its very roots. Image de-noising techniques [1] have been deployed at various levels of abstraction to make life easier for image and data scientists who rely solely on image quality for continuation of their important research. Hubble Space Telescope(HST) have been sending us photographs of planets, their moons, distant galaxies, birth and death of stars and its service has been top-notch in 99.99 per cent. of cases. Image processing and restoration from HST is a huge endeavour that is currently employing hundreds of brains across the planet [2]. The processing of such large volumes of data from such enormous sources requires lot of processing time and is very expensive. The area of hardware required to achieve excellent image quality increases proportionally with the complexity of the application the image will be put to use in. Keeping the sophistication of the process in mind, research is ongoing in computer science to develop algorithms for cheaper methods of processing such images. In this paper, a simple and elegant method has been described to process such sparse-coloured images which have been exacerbated by Gaussian noise in the communication channel. The algorithm, along with a brief comparison with the usual image filtering methods in practice, will initiate a radical new approach to image processing where images are particularly having specific colours in them and thus, the processing cost and area will come down manifold. Development on the grounds of this algorithmic model will lead to a whole new way of thinking in terms of redistributing noise instead of removing it and redistributing it to areas where the effect of the noise will be negligible but at the same time, the contrast of the image will rise and energy of the image will be conserved which can be utilised for more interesting purposes that are to be discussed in the future work section.

## II. LITERATURE REVIEW

The primary objective of this research is to identify the gaps in the current research going on in various fields of image processing directed towards de-noising images with AWGN and attempt to fill them for the greater good.

Several books by eminent authors explore the standard methods for image filtering and de-noising [3] along with different noise models. Most of these are very much in practice and saturation may be a problem which should be mitigated by more lateral methods to handle the noise in an image. De-noising image processes have been discussed in terms of comparison of image quality assessment, with respect to Peak Signal to Noise Ratio (PSNR), Human Visual System (HVS), Structural Similarity Index (SSIM) and Universal Image Quality Index (UIQI) [4]. These metrics are necessary for noise reduction and performance measurement and the method discussed in this paper have been tested on their grounds only.

Genetic algorithms have been developed for noise reduction in images. Kullback-Leibler divergence minimisation is employed in such methods [5]. However, it is mathematically more rigorous and computational cost is also higher.

Charged-couple device (CCD) noise is removed using non-linear filtering and enhancement [6]. The method in this paper resolves AWGN noise and is devoid of filtering operation which burns a lot of power. Minimal filtering may be required for testing and verification purposes.

In view of the above, having explored the various kinds of research covered in mitigating noise effect in images, it is always noise getting removed. Redistributing noise to parts of the image where they matter or affect less is clever way of handling the disturbance noise poses to the image and its applications. the image is hence effectively cleaned without having to actually filter or implement any large power or cost operation on it. Redistribution is motivated by conservation which in turn has come from the growing energy-resource concerns in the current world scenario. Therefore, not only this method proposes noise mitigation but also a cleaner and compact way of looking at de-noising an image in a manner

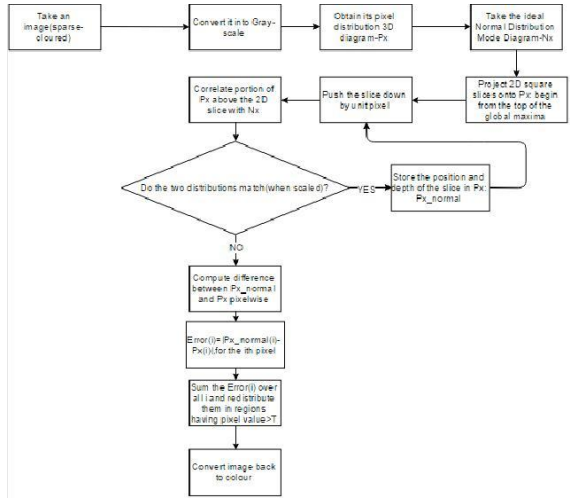

Fig. 1: Flow Chart for Selective De-noising Algorithm

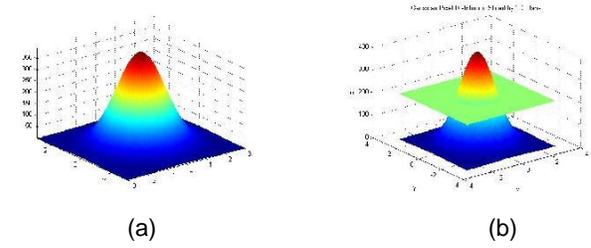

(a)  (b)

Fig. 2: (a) Gaussian 3D Pixel Distribution P (x, y, z) (b) 2D Slice at depth d

that it serves both purposes-increasing image contrast and maintaining image integrity.

### III. Approach

Spatial and frequency domain filtering of images are nowadays extensively used for noise reduction. However, the mechanism of selective(or intelligent) redistribution of noise elements is a technique which can solve the same noise 'reduction' problem from a more energy-efficient perspective. The flowchart of the algorithm very briefly describes the direction of operation (Figure:1). Details of the process have been discussed in the subsequent subsections.

#### A. Conversion Rules for RGB to Gray-scale

If each colour pixel is described by a triplet (R; G; B) of intensities for red, green, and blue components, they can be mapped to a single number giving a gray-scale value.

#### B. Normal Distribution Detection in Image

A three dimensional representation of the pixel magnitude distribution is computed and shown on a plot with larger pixel values depicted as taller stacks. The fact that an image is a 2D matrix $A_{mxn}$ consisting of pixel values as its (i, j) elements make the mathematical detection of a normal distribution detection simpler in the given pixel distribution. As the image has been affected by Gaussian white noise, the additive nature of the noise will cause the envelope of the distribution of pixel stacks to be Gaussian distribution. However, the challenge lies in finding the base of that distribution in the 3D diagram. To find the base, we project a 2D slice (Figure:2) which is a square(for convenience) cutting the 3D model of pixel distribution P (x, y, z), where

$$z = A(i, j); i = x; j = y;$$

horizontally, beginning at the topmost peak of P (x, y, z). The following equation summarises the initial state of the variables (variables include-$h_{initial}$, d, etc.):

$$h_{initial} = max(pixel(k))$$

where pixel(k) = A(i, j);

where $h_{initial}$ is the maximum depth of the pixel distribution. With every iteration of unit pixel, the slice is pushed down, i.e.

$$d = d + 1; d_{initial} = 0$$

and the 3D portion of P (x, y, z) above the 2D slice P" (x, y, z) is compared with a scaled version(to match the dimensions) of an ideal 3D Gaussian model G(x, y, z). In case of a match, the depth d of the 2D slice in P (x, y, z) is stored in $d_{temp}$, forcing next iteration. When the match fails to occur for the first time, $d_{temp}$, which forms the basis, effectively the platform on which the additive white Gaussian noise rests atop the actual pixel distribution $P_{actual}(x, y, z)$, is used as a reference mark for further computations.

#### C. Error Calculation and Accumulation

The error is calculated pixel wise by calculating the absolute difference between actual pixel distribution and the AWGN envelope estimated.

$$Err(i,j) = |P_{actual}(i,j) - P_{normal}(i,j)|$$

where $P_{actual}(i,j) = |z_{P(x,y,z)} - d_{temp}|, i = x, j = y$
and $P_{normal}(i,j) = |z_{G(x,y,z)} - d_{temp}|, i = x, j = y$.
Accumulation of the error is done by:

$$Error_{acc} = \sum_{1}^{m}\sum_{1}^{n} Err(i,j)$$

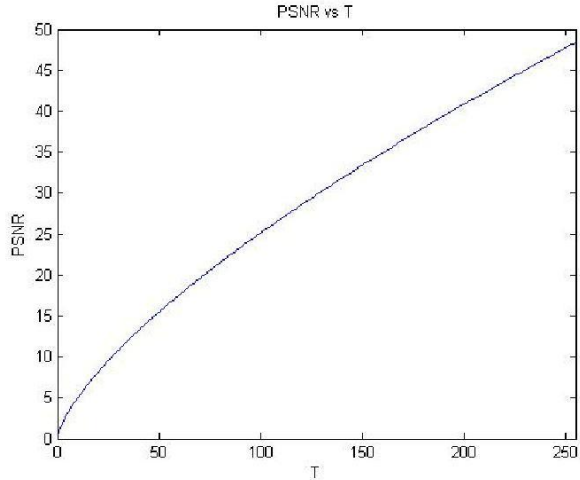

Fig. 3: Variation of PSNR Performance with T (Y-axis normalised to 50.0)

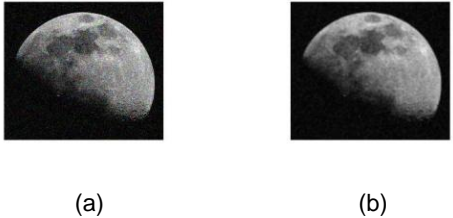

(a)  (b)

Fig. 4: (a) Moon with Noise
(b)Moon After Noise Removal

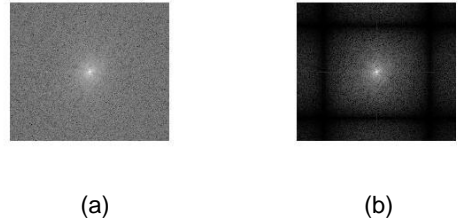

(a)  (b)

Fig. 5: (a) FFT of Moon with Noise
(b) FFT of Moon Post Noise Removal

### C. Redistribution of the Accumulated Error

A certain threshold T is defined in the range [0, 255] to ascertain the redistribution of the $Error_{acc}$ to various parts of the image. Some value is subtracted from the accumulated error and imparted to increase the pixel value of the actual pixel distribution where $P_{actual}(i,j)$ is greater than a certain threshold value, say T . This is done to ensure that whiter regions of the gray-scale image become more white(pixel value being 255), allowing negligible visual effect of noise. Spectral components in such high pixel value regions will also be of much less concern and importance-the whiter regions. The choice of the threshold parameter T determines the contrast and the PSNR (Peak Signal to Noise Ratio). Thereby, effectively, the noise gets rehabilitated to a new area where its effect will be infinitesimal to the analysis of the image's more important regions. This is maximally effective in case of sparse-coloured astronomical images where less number of distinct-coloured regions are there and thereby, larger area of 'white' patches are present.

### IV. TESTING AND RESULTS

This selective noise mitigation scheme is tested in a MATLAB Simulation environment. The following interesting observations came out from the simulations:

The choice of T, as earlier mentioned, affects the PSNR and the contrast of the image after processing. Selective redistribution of the noise leads to an excellent result with large PSNR, depicting higher quality of the image. Larger the value of T, closer it is to the white (255) component, higher is the PSNR (calculated with respect to the crucial or desired parts of the image) and hence better the result as white absorbs the extra pixel values added to it, without any marked change in picture quality, whereby on the other hand, reduces noise disturbance in the crucial parts of the image (Figure:3).

It may happen as in this case, that $Error_{acc}$ is not completely exhausted after redistribution of the pixel values. That will still be better than the ordinary filtering operations where the energy of the original image is reduced more owing to complete removal of the noise pixels whereas, here, the larger portion (about 75 per cent.) of it is sustained.

The original image with noise has been de-noised completely in agreement with our proposed theory and the simulation results have an excellent correlation with the algorithm's expectations (Figure:4).

### V. CONCLUSION AND FUTURE WORK

This paper presents an out-of-the-box approach to image processing where noise instead of being exterminated, is shifted to certain locations within the same image with minimum loss of average image energy. The selectivity of this entire mechanism leads to the flexibility in choosing the T value to our own convenience and interest of application. Owing to the fact that the original image pixel distribution estimate is required for error calculation, this methods will find wider application in sparse-coloured image processing-in vivid astrophysical studies- where the gradient in the images will be low, leading to similar pixel values and hence, the Gaussian envelope's lower bound or base-platform can be easily estimated. This energy-efficient method eliminates the use of power-consuming ADCs, DACs and Band-Pass filters, and maintains the integrity of the original image with reduced hardware. The residual noise energy in the image can be utilised further to increase image contrast for better analysis.

Artificial intelligence can be put to use to classify the sparse-coloured images in finer ways for better PSNR and energy-sustainability. As we move into a generation of more advanced technology, the probabilistic nature of the white Gaussian noise will be taken care of in a more subtle and elegant manner, so much so as to foster the use of this method in other fields of image and signal processing. The hypothetical nature of this solution to the noise problem in images may thrust further research into this particular avenue of thoughts for improvement in noise and distortion handling in digital image processing.